# Identifying the Relevant Nodes Without Learning the Model

**Jose M. Peña**
Linköping University
58183 Linköping, Sweden

**Roland Nilsson**
Linköping University
58183 Linköping, Sweden

**Johan Björkegren**
Karolinska Institutet
17177 Stockholm, Sweden

**Jesper Tegnér**
Linköping University
58183 Linköping, Sweden

## Abstract

We propose a method to identify all the nodes that are relevant to compute all the conditional probability distributions for a given set of nodes. Our method is simple, efficient, consistent, and does not require learning a Bayesian network first. Therefore, our method can be applied to high-dimensional databases, e.g. gene expression databases.

## 1 INTRODUCTION

As part of our project on atherosclerosis gene expression data analysis, we want to learn a Bayesian network (BN) to answer any query about the state of certain atherosclerosis genes given the state of any other set of genes. If $\mathbf{U}$ denotes all the nodes (genes) and $\mathbf{T} \subseteq \mathbf{U}$ denotes the target nodes (atherosclerosis genes), then we want to learn a BN to answer any query of the form $p(\mathbf{T}|\mathbf{Z} = \mathbf{z})$ with $\mathbf{Z} \subseteq \mathbf{U} \setminus \mathbf{T}$. Unfortunately, learning a BN for $\mathbf{U}$ is impossible with our resources because $\mathbf{U}$ contains thousands of nodes. However, we do not really need to learn a BN for $\mathbf{U}$ to be able to answer any query about $\mathbf{T}$: It suffices to learn a BN for $\mathbf{U} \setminus \mathbf{I}$ where $\mathbf{I}$, the irrelevant nodes, is a maximal subset of $\mathbf{U} \setminus \mathbf{T}$ such that $\mathbf{I} \perp\!\!\!\perp \mathbf{T}|\mathbf{Z}$ for all $\mathbf{Z} \subseteq \mathbf{U} \setminus \mathbf{T} \setminus \mathbf{I}$. We prove that $\mathbf{I}$ is unique and, thus, that $\mathbf{U} \setminus \mathbf{I}$ is the optimal domain to learn the BN, because $\mathbf{R} = \mathbf{U} \setminus \mathbf{T} \setminus \mathbf{I}$ is the (unique) minimal subset of $\mathbf{U} \setminus \mathbf{T}$ that contains all the nodes that are relevant to answer all the queries about $\mathbf{T}$.

We propose the following method to identify $\mathbf{R}$: $R \in \mathbf{R}$ iff there exists a sequence of nodes starting with $R$ and ending with some $T \in \mathbf{T}$ such that every two consecutive nodes in the sequence are marginally dependent. We prove that the method is consistent under the assumptions of strict positivity, composition and weak transitivity. We argue that these assumptions are not too restrictive. It is worth noting that our method is efficient in terms of both runtime and data requirements and, thus, can be applied to high-dimensional domains. We believe that our method can be helpful to those working with such domains, where identifying the optimal domain before learning the BN can reduce costs drastically.

Although, to our knowledge, the problem addressed in this paper has not been studied before, there do exist papers that address closely related problems. Geiger et al. (1990) and Shachter (1988, 1990, 1998) show how to identify in a BN structure all the nodes whose BN parameters are needed to answer a particular query. Lin and Druzdzel (1997) show how to identify some nodes that can be removed from a BN without affecting the answer to a particular query. Mahoney and Laskey (1998) propose an algorithm based on the work by Lin and Druzdzel (1997) to construct, from a set of BN fragments, a minimal BN to answer a particular query. Madsen and Jensen (1998) show how to identify some operations in the junction tree of a BN that can be skipped without affecting the answer to a particular query. Two are the main differences between these works and our contribution. First, they focus on a single query while we focus on all the queries about the target nodes. Second, they require learning a BN structure first while we do not. Before going into the details of our contribution, we review some key concepts in the following section.

## 2 PRELIMINARIES

The following definitions and results can be found in most books on Bayesian networks, e.g. Pearl (1988) and Studený (2005). Let $\mathbf{U}$ denote a non-empty finite set of random variables. A Bayesian network (BN) for $\mathbf{U}$ is a pair $(G, \theta)$ where $G$, the structure, is a directed and acyclic graph (DAG) whose nodes correspond to the random variables in $\mathbf{U}$ and $\theta$, the parameters, are parameters specifying a probability distribution for each $X \in \mathbf{U}$ given its parents in $G$,

$p(X|Pa(X))$. A BN $(G, \theta)$ represents a probability distribution for $\mathbf{U}$, $p(\mathbf{U})$, through the factorization $p(\mathbf{U}) = \prod_{X \in \mathbf{U}} p(X|Pa(X))$. Therefore, it is clear that a BN for $\mathbf{U}$ can answer any query $p(\mathbf{T}|\mathbf{Z} = \mathbf{z})$ with $\mathbf{T} \subseteq \mathbf{U}$ and $\mathbf{Z} \subseteq \mathbf{U} \setminus \mathbf{T}$. Hereinafter, all the probability distributions and DAGs are defined over $\mathbf{U}$, unless otherwise stated.

Let $\mathbf{X}$, $\mathbf{Y}$ and $\mathbf{Z}$ denote three mutually disjoint subsets of $\mathbf{U}$. Let $\mathbf{X} \perp\!\!\!\perp \mathbf{Y} | \mathbf{Z}$ denote that $\mathbf{X}$ is independent of $\mathbf{Y}$ given $\mathbf{Z}$ in a probability distribution $p$. Let $sep(\mathbf{X}, \mathbf{Y} | \mathbf{Z})$ denote that $\mathbf{X}$ is separated from $\mathbf{Y}$ by $\mathbf{Z}$ in a graph $G$. If $G$ is a DAG, then $sep(\mathbf{X}, \mathbf{Y} | \mathbf{Z})$ is true when for every undirected path in $G$ between a node in $\mathbf{X}$ and a node in $\mathbf{Y}$ there exists a node $Z$ in the path such that either (i) $Z$ does not have two parents in the path and $Z \in \mathbf{Z}$, or (ii) $Z$ has two parents in the path and neither $Z$ nor any of its descendants in $G$ is in $\mathbf{Z}$. On the other hand, if $G$ is an undirected graph (UG), then $sep(\mathbf{X}, \mathbf{Y} | \mathbf{Z})$ is true when for every path in $G$ between a node in $\mathbf{X}$ and a node in $\mathbf{Y}$ there exists some $Z \in \mathbf{Z}$ in the path. A probability distribution $p$ is faithful to a DAG or UG $G$ when $\mathbf{X} \perp\!\!\!\perp \mathbf{Y} | \mathbf{Z}$ iff $sep(\mathbf{X}, \mathbf{Y} | \mathbf{Z})$. $G$ is an independence map of $p$ when $\mathbf{X} \perp\!\!\!\perp \mathbf{Y} | \mathbf{Z}$ if $sep(\mathbf{X}, \mathbf{Y} | \mathbf{Z})$. $G$ is a minimal independence map of $p$ when removing any edge from $G$ makes it cease to be an independence map of $p$. If $p$ is strictly positive, then it has a unique minimal undirected independence map $G$, and it can be built via the edge exclusion algorithm: Two nodes $X$ and $Y$ are adjacent in $G$ iff $X \not\!\perp\!\!\!\perp Y | \mathbf{U} \setminus \{X, Y\}$. Alternatively, $G$ can be built via the Markov boundary algorithm: Two nodes $X$ and $Y$ are adjacent in $G$ iff $Y$ belongs to the Markov boundary of $X$. The Markov boundary of $X$ is the set $\mathbf{X} \subseteq \mathbf{U} \setminus \{X\}$ such that (i) $X \perp\!\!\!\perp \mathbf{U} \setminus \mathbf{X} \setminus \{X\} | \mathbf{X}$, and (ii) no proper subset of $\mathbf{X}$ satisfies (i).

Let $\mathbf{X}$, $\mathbf{Y}$, $\mathbf{Z}$ and $\mathbf{W}$ denote four mutually disjoint subsets of $\mathbf{U}$. Any probability distribution $p$ satisfies the following four properties: Symmetry $\mathbf{X} \perp\!\!\!\perp \mathbf{Y} | \mathbf{Z} \Rightarrow \mathbf{Y} \perp\!\!\!\perp \mathbf{X} | \mathbf{Z}$, decomposition $\mathbf{X} \perp\!\!\!\perp \mathbf{Y} \cup \mathbf{W} | \mathbf{Z} \Rightarrow \mathbf{X} \perp\!\!\!\perp \mathbf{Y} | \mathbf{Z}$, weak union $\mathbf{X} \perp\!\!\!\perp \mathbf{Y} \cup \mathbf{W} | \mathbf{Z} \Rightarrow \mathbf{X} \perp\!\!\!\perp \mathbf{Y} | \mathbf{Z} \cup \mathbf{W}$, and contraction $\mathbf{X} \perp\!\!\!\perp \mathbf{Y} | \mathbf{Z} \cup \mathbf{W} \wedge \mathbf{X} \perp\!\!\!\perp \mathbf{W} | \mathbf{Z} \Rightarrow \mathbf{X} \perp\!\!\!\perp \mathbf{Y} \cup \mathbf{W} | \mathbf{Z}$. If $p$ is strictly positive, then it satisfies the intersection property $\mathbf{X} \perp\!\!\!\perp \mathbf{Y} | \mathbf{Z} \cup \mathbf{W} \wedge \mathbf{X} \perp\!\!\!\perp \mathbf{W} | \mathbf{Z} \cup \mathbf{Y} \Rightarrow \mathbf{X} \perp\!\!\!\perp \mathbf{Y} \cup \mathbf{W} | \mathbf{Z}$. If $p$ is DAG-faithful or UG-faithful, then it satisfies the following two properties: Composition $\mathbf{X} \perp\!\!\!\perp \mathbf{Y} | \mathbf{Z} \wedge \mathbf{X} \perp\!\!\!\perp \mathbf{W} | \mathbf{Z} \Rightarrow \mathbf{X} \perp\!\!\!\perp \mathbf{Y} \cup \mathbf{W} | \mathbf{Z}$, and weak transitivity $\mathbf{X} \perp\!\!\!\perp \mathbf{Y} | \mathbf{Z} \wedge \mathbf{X} \perp\!\!\!\perp \mathbf{Y} | \mathbf{Z} \cup \{W\} \Rightarrow \mathbf{X} \perp\!\!\!\perp \{W\} | \mathbf{Z} \vee \{W\} \perp\!\!\!\perp \mathbf{Y} | \mathbf{Z}$ with $W \in \mathbf{U} \setminus \mathbf{X} \setminus \mathbf{Y} \setminus \mathbf{Z}$.

## 3 IDENTIFYING THE RELEVANT NODES

We say that $\mathbf{X} \subseteq \mathbf{U} \setminus \mathbf{T}$ is irrelevant to answer any query about $\mathbf{T} \subseteq \mathbf{U}$ when $\mathbf{X} \perp\!\!\!\perp \mathbf{T} | \mathbf{Z}$ for all $\mathbf{Z} \subseteq \mathbf{U} \setminus \mathbf{T} \setminus \mathbf{X}$, because $p(\mathbf{T}|\mathbf{Z}, \mathbf{Z}') = p(\mathbf{T}|\mathbf{Z})$ for all $\mathbf{Z} \subseteq \mathbf{U} \setminus \mathbf{T} \setminus \mathbf{X}$ and $\mathbf{Z}' \subseteq \mathbf{X}$ due to decomposition. Therefore, we do not really need to learn a BN for $\mathbf{U}$ to be able to answer any query about $\mathbf{T}$: It suffices to learn a BN for $\mathbf{U} \setminus \mathbf{X}$. The following theorem characterizes all the sets of nodes that are irrelevant.

**Theorem 1** *Let* $\mathbf{I} = \{X_1, \ldots, X_n\}$ *denote all the nodes in* $\mathbf{U} \setminus \mathbf{T}$ *such that, for all $i$,* $X_i \perp\!\!\!\perp \mathbf{T} | \mathbf{Z}$ *for all* $\mathbf{Z} \subseteq \mathbf{U} \setminus \mathbf{T} \setminus \{X_i\}$. *Then,* $\mathbf{I}$ *is the (unique) maximal subset of* $\mathbf{U} \setminus \mathbf{T}$ *such that* $\mathbf{I} \perp\!\!\!\perp \mathbf{T} | \mathbf{Z}$ *for all* $\mathbf{Z} \subseteq \mathbf{U} \setminus \mathbf{T} \setminus \mathbf{I}$.

**Proof:** Let $\mathbf{Z} \subseteq \mathbf{U} \setminus \mathbf{T} \setminus \mathbf{I}$. Since $X_1 \perp\!\!\!\perp \mathbf{T} | \mathbf{Z}$ and $X_2 \perp\!\!\!\perp \mathbf{T} | \mathbf{Z} \cup \{X_1\}$, then $\{X_1, X_2\} \perp\!\!\!\perp \mathbf{T} | \mathbf{Z}$ due to contraction. This together with $X_3 \perp\!\!\!\perp \mathbf{T} | \mathbf{Z} \cup \{X_1, X_2\}$ implies $\{X_1, X_2, X_3\} \perp\!\!\!\perp \mathbf{T} | \mathbf{Z}$ due to contraction again. Continuing this process for the rest of the nodes in $\mathbf{I}$ proves that $\mathbf{I} \perp\!\!\!\perp \mathbf{T} | \mathbf{Z}$.

Let us assume that there exists some $\mathbf{I}' \subseteq \mathbf{U}$ such that $\mathbf{I}' \setminus \mathbf{I} \neq \emptyset$ and $\mathbf{I}' \perp\!\!\!\perp \mathbf{T} | \mathbf{Z}$ for all $\mathbf{Z} \subseteq \mathbf{U} \setminus \mathbf{T} \setminus \mathbf{I}'$. Let $X \in \mathbf{I}' \setminus \mathbf{I}$. Then, $X \perp\!\!\!\perp \mathbf{T} | \mathbf{Z}$ for all $\mathbf{Z} \subseteq \mathbf{U} \setminus \mathbf{T} \setminus \{X\}$ due to decomposition and weak union. This is a contradiction because $X \notin \mathbf{I}$. Consequently, $\mathbf{I}$ is the (unique) maximal subset of $\mathbf{U}$ such that $\mathbf{I} \perp\!\!\!\perp \mathbf{T} | \mathbf{Z}$ for all $\mathbf{Z} \subseteq \mathbf{U} \setminus \mathbf{T} \setminus \mathbf{I}$. □

It follows from the theorem above that a set of nodes is irrelevant iff it is a subset of $\mathbf{I}$ due to decomposition and weak union. Therefore, $\mathbf{U} \setminus \mathbf{I}$ is the optimal domain to learn the BN, because $\mathbf{R} = \mathbf{U} \setminus \mathbf{T} \setminus \mathbf{I}$ is the (unique) minimal subset of $\mathbf{U} \setminus \mathbf{T}$ that contains all the nodes that are relevant to answer all the queries about $\mathbf{T}$. The theorem above characterizes $\mathbf{R}$ as $X \in \mathbf{R}$ iff $X \not\!\perp\!\!\!\perp \mathbf{T} | \mathbf{Z}$ for some $\mathbf{Z} \subseteq \mathbf{U} \setminus \mathbf{T} \setminus \{X\}$. Unfortunately, this is not a practical characterization due to the potentially huge number of conditioning sets to consider. The following theorem gives a more practical characterization of $\mathbf{R}$.

**Theorem 2** *Let $p$ be a strictly positive probability distribution satisfying weak transitivity. Then,* $X \in \mathbf{R}$ *iff there exists a path between $X$ and some* $T \in \mathbf{T}$ *in the minimal undirected independence map $G$ of $p$.*

**Proof:** If there exists no path for $X$ like the one described in the theorem, then $X \perp\!\!\!\perp \mathbf{T} | \mathbf{Z}$ for all $\mathbf{Z} \subseteq \mathbf{U} \setminus \mathbf{T} \setminus \{X\}$ because $G$ is an undirected independence map of $p$. Consequently, $X \notin \mathbf{R}$ by Theorem 1.

Let $X_1, \ldots, X_n$ with $X_i \in \mathbf{U} \setminus \mathbf{T}$ for all $i < n$ and $X_n \in \mathbf{T}$ denote the sequence of nodes in the shortest path in $G$ between $X_1$ and a node in $\mathbf{T}$. Since $G$ is the minimal undirected independence map of $p$, then

$$X_i \not\!\perp\!\!\!\perp X_j | \mathbf{U} \setminus \{X_i, X_j\} \qquad (1)$$

iff $X_i$ and $X_j$ are consecutive in the sequence. We prove that $X_1 \not\!\perp\!\!\!\perp X_n | \mathbf{U} \setminus \{X_1, \ldots, X_n\}$, which implies

that $X_1 \not\!\perp\!\!\!\perp \mathbf{T} | \mathbf{U} \setminus \mathbf{T} \setminus \{X_1, \ldots, X_{n-1}\}$ due to weak union and, thus, that $X_1 \in \mathbf{R}$ by Theorem 1. If $n = 2$, then this is true by equation (1). We now prove it for $n > 2$. We start by proving that $X_i \not\!\perp\!\!\!\perp X_j | \mathbf{U} \setminus \{X_i, X_j, X_k\}$ for all $X_i$ and $X_j$ that are consecutive in the sequence. Let us assume that $i < j < k$. The proof is analogous for $k < i < j$. By equation (1),

$$X_i \not\!\perp\!\!\!\perp X_j | \mathbf{U} \setminus \{X_i, X_j\} \tag{2}$$

and
$$X_i \perp\!\!\!\perp X_k | \mathbf{U} \setminus \{X_i, X_k\}. \tag{3}$$

Let us assume that
$$X_i \perp\!\!\!\perp X_j | \mathbf{U} \setminus \{X_i, X_j, X_k\}. \tag{4}$$

Then, $X_i \perp\!\!\!\perp \{X_j, X_k\} | \mathbf{U} \setminus \{X_i, X_j, X_k\}$ due to contraction on equations (3) and (4) and, thus, $X_i \perp\!\!\!\perp X_j | \mathbf{U} \setminus \{X_i, X_j\}$ due to weak union. This contradicts equation (2) and, thus,

$$X_i \not\!\perp\!\!\!\perp X_j | \mathbf{U} \setminus \{X_i, X_j, X_k\}. \tag{5}$$

We now prove that $X_i \not\!\perp\!\!\!\perp X_k | \mathbf{U} \setminus \{X_i, X_j, X_k\}$ for all $X_i$, $X_j$ and $X_k$ that are consecutive in the sequence. By equation (5), $X_i \not\!\perp\!\!\!\perp X_j | \mathbf{U} \setminus \{X_i, X_j, X_k\}$ and $X_j \not\!\perp\!\!\!\perp X_k | \mathbf{U} \setminus \{X_i, X_j, X_k\}$ and, thus, $X_i \not\!\perp\!\!\!\perp X_k | \mathbf{U} \setminus \{X_i, X_j, X_k\}$ or $X_i \not\!\perp\!\!\!\perp X_k | \mathbf{U} \setminus \{X_i, X_k\}$ due to weak transitivity. Since the latter contradicts equation (1), we conclude that

$$X_i \not\!\perp\!\!\!\perp X_k | \mathbf{U} \setminus \{X_i, X_j, X_k\}. \tag{6}$$

Finally, we prove that $X_i \perp\!\!\!\perp X_j | \mathbf{U} \setminus \{X_i, X_j, X_k\}$ for all $X_i$, $X_j$ and $X_k$ such that neither the first two nor the last two are consecutive in the sequence. By equation (1), $X_i \perp\!\!\!\perp X_j | \mathbf{U} \setminus \{X_i, X_j\}$ and $X_j \perp\!\!\!\perp X_k | \mathbf{U} \setminus \{X_j, X_k\}$. Then,

$$X_i \perp\!\!\!\perp X_j | \mathbf{U} \setminus \{X_i, X_j, X_k\} \tag{7}$$

due to intersection and decomposition.

It can be seen from equations (5), (6) and (7) that the sequence $X_1, X_3, \ldots, X_n$ satisfies equation (1) replacing $\mathbf{U}$ by $\mathbf{U} \setminus \{X_2\}$: Equations (5) and (6) ensure that every two consecutive nodes are dependent, while equation (7) ensures that every two non-consecutive nodes are independent. Therefore, we can repeat the calculations above for the sequence $X_1, X_3, \ldots, X_n$ replacing $\mathbf{U}$ by $\mathbf{U} \setminus \{X_2\}$. This allows us to successively remove the nodes $X_2, \ldots, X_{n-1}$ from the sequence $X_1, \ldots, X_n$ and conclude that the sequence $X_1, X_n$ satisfies equation (1) replacing $\mathbf{U}$ by $\mathbf{U} \setminus \{X_2, \ldots, X_{n-1}\}$. Then, $X_1 \not\!\perp\!\!\!\perp X_n | \mathbf{U} \setminus \{X_1, \ldots, X_n\}$. □

In the theorem above, we do not really need to learn $G$: It suffices to identify the nodes in the connected components of $G$ that include some $T \in \mathbf{T}$. These nodes can be identified as follows. First, initialize $\mathbf{R}$ with $\mathbf{T}$. Second, repeat the following step while possible: For each node in $\mathbf{R}$ that has not been considered before, find all the nodes in $\mathbf{U} \setminus \mathbf{R}$ that are adjacent to it in $G$ and add them to $\mathbf{R}$. Finally, remove $\mathbf{T}$ from $\mathbf{R}$. The second step can be solved with the help of the edge exclusion algorithm or the Markov boundary algorithm. The conditioning set in every independence test that the edge exclusion algorithm performs is of size $|\mathbf{U}|-2$. On the other hand, the largest conditioning set in the tests that the Markov boundary algorithm performs is at least of the size of the largest Markov boundary in the connected components.[1] Therefore, both algorithms can require a large learning database to return the true adjacent nodes with high probability, because the conditioning sets in some of the tests can be quite large. This is a problem if only a small learning database is available as, for instance, in gene expression data analysis where collecting data is expensive. The following theorem gives a characterization of $\mathbf{R}$ that only tests for marginal independence and, thus, the method to identify $\mathbf{R}$ it gives rise to requires minimal learning data.

**Theorem 3** *Let $p$ be a strictly positive probability distribution satisfying composition and weak transitivity. Then, $X_1 \in \mathbf{R}$ iff there exists a sequence $X_1, \ldots, X_n$ with $X_i \in \mathbf{U} \setminus \mathbf{T}$ for all $i < n$, $X_n \in \mathbf{T}$, and $X_i \not\!\perp\!\!\!\perp X_{i+1} | \emptyset$ for all $i$.*

**Proof:** Let $\mathbf{X}$ denote all the nodes in $\mathbf{U} \setminus \mathbf{T}$ for which there exists a sequence like the one described in the theorem. Let $\mathbf{Y} = \mathbf{U} \setminus \mathbf{T} \setminus \mathbf{X}$. Let $\mathbf{W}$ denote all the nodes in $\mathbf{U} \setminus \mathbf{T}$ from which there exists a path to some $T \in \mathbf{T}$ in the minimal undirected independence map of $p$. If $X \in \mathbf{X}$ then all the nodes in its sequence except the last one must be in $\mathbf{W}$, otherwise there is a contradiction because two adjacent nodes in the sequence are marginally independent in $p$. Then, $\mathbf{X} \subseteq \mathbf{W}$ and, thus, $\mathbf{X} \subseteq \mathbf{R}$ because $\mathbf{R} = \mathbf{W}$ by Theorem 2.

Moreover, $X \perp\!\!\!\perp Y | \emptyset$ and $Y \perp\!\!\!\perp T | \emptyset$ for all $X \in \mathbf{X}$, $Y \in \mathbf{Y}$ and $T \in \mathbf{T}$, otherwise there is a contradiction because there exists a sequence for $Y$ like the one described in the theorem. Then, $\mathbf{Y} \perp\!\!\!\perp \mathbf{X} \cup \mathbf{T} | \emptyset$ due to composition and, thus, $\mathbf{Y} \perp\!\!\!\perp \mathbf{T} | \mathbf{Z}$ for all $\mathbf{Z} \subseteq \mathbf{U} \setminus \mathbf{T} \setminus \mathbf{Y}$ due to decomposition and weak union. Consequently, $\mathbf{Y} \subseteq \mathbf{I}$ by Theorem 1 and, thus, $\mathbf{R} \subseteq \mathbf{X}$. □

In the method to identify $\mathbf{R}$ that follows from the theorem above, we do not really need to perform all the $|\mathbf{U}|(|\mathbf{U}|-1)/2$ marginal independence tests if we adopt the following implementation. First, initialize $\mathbf{R}$ with $\mathbf{T}$. Second, repeat the following step while possible: For each node in $\mathbf{R}$ that has not been considered before, find all the nodes in $\mathbf{U} \setminus \mathbf{R}$ that are marginally dependent on it and add them to $\mathbf{R}$. Finally, remove $\mathbf{T}$ from $\mathbf{R}$. Therefore, this implementation is efficient in terms of both runtime and data requirements and, thus, can be applied to high-dimensional domains, where identifying the optimal domain before learning the BN can reduce costs drastically. It is also worth noting that the irrelevant nodes are not necessarily mutually independent, i.e. they can have an arbitrary dependence structure.

---

[1] We assume that the Markov boundary of a node is obtained via the incremental association Markov boundary algorithm (Tsamardinos et al. 2003) which, to our knowledge, is the only existing algorithm that satisfies our requirements of scalability and consistency when assuming composition (Peña et al. 2006a).

It goes without saying that there is no guarantee that $\mathbf{U} \setminus \mathbf{I}$ will not still be too large to learn a BN. If this is the case, then one may have to reduce $\mathbf{T}$. Another solution may be to consider only the queries about $\mathbf{T}$ where the conditioning set includes the context nodes $\mathbf{C} \subseteq \mathbf{U} \setminus \mathbf{T}$. In other words, the BN to be learnt should be able to answer any query $p(\mathbf{T}|\mathbf{C}=\mathbf{c}, \mathbf{Z}=\mathbf{z})$ with $\mathbf{Z} \subseteq \mathbf{U} \setminus \mathbf{T} \setminus \mathbf{C}$ but not the rest. Repeating our reasoning above, it suffices to learn a BN for $\mathbf{U} \setminus \mathbf{I}(\mathbf{C})$ where $\mathbf{I}(\mathbf{C})$ is a maximal subset of $\mathbf{U} \setminus \mathbf{T} \setminus \mathbf{C}$ such that $\mathbf{I}(\mathbf{C}) \perp\!\!\!\perp \mathbf{T}|\mathbf{C} \cup \mathbf{Z}$ for all $\mathbf{Z} \subseteq \mathbf{U} \setminus \mathbf{T} \setminus \mathbf{C} \setminus \mathbf{I}(\mathbf{C})$. The following theorem shows that $\mathbf{I}(\mathbf{C})$ can be obtained by applying Theorem 3 to $p(\mathbf{U}\setminus\mathbf{C}|\mathbf{C}=\mathbf{c})$ for any $\mathbf{c}$, under the assumption that $p(\mathbf{U} \setminus \mathbf{C}|\mathbf{C}=\mathbf{c})$ has the same independencies for all $\mathbf{c}$. We discuss this assumption in the next section.

**Theorem 4** *Let $p$ be a strictly positive probability distribution satisfying composition and weak transitivity, and such that $p(\mathbf{U} \setminus \mathbf{C}|\mathbf{C}=\mathbf{c})$ has the same independencies for all $\mathbf{c}$. Then, the result of applying Theorem 3 to $p(\mathbf{U} \setminus \mathbf{C}|\mathbf{C}=\mathbf{c})$ for any $\mathbf{c}$ is $\mathbf{R}(\mathbf{C}) = \mathbf{U} \setminus \mathbf{T} \setminus \mathbf{C} \setminus \mathbf{I}(\mathbf{C})$. Equivalently, $X_1 \in \mathbf{R}(\mathbf{C})$ iff there exists a sequence $X_1, \ldots, X_n$ with $X_i \in \mathbf{U} \setminus \mathbf{T} \setminus \mathbf{C}$ for all $i < n$, $X_n \in \mathbf{T}$, and $X_i \not\perp\!\!\!\perp X_{i+1}|\mathbf{C}$ for all $i$.*

**Proof:** Let $\mathbf{X}$, $\mathbf{Y}$ and $\mathbf{Z}$ denote three mutually disjoint subsets of $\mathbf{U} \setminus \mathbf{C}$. Since $p(\mathbf{U} \setminus \mathbf{C}|\mathbf{C}=\mathbf{c})$ has the same independencies for all $\mathbf{c}$ then, for any $\mathbf{c}$, $\mathbf{X} \perp\!\!\!\perp \mathbf{Y}|\mathbf{Z}$ in $p(\mathbf{U} \setminus \mathbf{C}|\mathbf{C}=\mathbf{c})$ iff $\mathbf{X} \perp\!\!\!\perp \mathbf{Y}|(\mathbf{Z} \cup \mathbf{C})$ in $p$. This has two implications. First, Theorem 3 can be applied to $p(\mathbf{U} \setminus \mathbf{C}|\mathbf{C}=\mathbf{c})$ for any $\mathbf{c}$ because it satisfies the strict positivity, composition and weak transitivity properties since $p$ satisfies them. Second, the result of applying Theorem 3 to $p(\mathbf{U} \setminus \mathbf{C}|\mathbf{C}=\mathbf{c})$ for any $\mathbf{c}$ is $\mathbf{R}(\mathbf{C})$. However, $X_i \not\perp\!\!\!\perp X_{i+1}|\emptyset$ in $p(\mathbf{U} \setminus \mathbf{C}|\mathbf{C}=\mathbf{c})$ iff $X_i \not\perp\!\!\!\perp X_{i+1}|\mathbf{C}$ in $p$ for all $i$. □

We note that the theorem above implies that $\mathbf{I}(\mathbf{C})$ is unique. It is also worth noting that we have not claimed that $\mathbf{U} \setminus \mathbf{I}(\mathbf{C})$ is the optimal domain to learn the BN. The reason is that it may not be minimal, e.g. if $\mathbf{C}$ includes some nodes from $\mathbf{I}$. To identify the optimal domain to learn the BN, $\mathbf{C}$ has to be purged as follows before running Theorem 4: Find some $X \in \mathbf{C}$ such that $X \notin \mathbf{R}(\mathbf{C} \setminus \{X\})$, remove $X$ from $\mathbf{C}$, and continue purging the resulting set. We prove that purging a node never makes an irrelevant node (including those purged before) become relevant. It suffices to prove that if $Y \notin \mathbf{R}(\mathbf{C})$, $X \in \mathbf{C}$, and $X \notin \mathbf{R}(\mathbf{C} \setminus \{X\})$, then $Y \notin \mathbf{R}(\mathbf{C} \setminus \{X\})$. Let $\mathbf{Z} \subseteq \mathbf{U} \setminus \mathbf{T} \setminus (\mathbf{C} \setminus \{X\}) \setminus \{Y\}$. If $X \in \mathbf{Z}$ then $Y \perp\!\!\!\perp \mathbf{T}|(\mathbf{C} \setminus \{X\}) \cup \mathbf{Z}$ because $Y \notin \mathbf{R}(\mathbf{C})$. On the other hand, if $X \notin \mathbf{Z}$ then $Y \perp\!\!\!\perp \mathbf{T}|\mathbf{C} \cup \mathbf{Z}$ because $Y \notin \mathbf{R}(\mathbf{C})$ and $X \perp\!\!\!\perp \mathbf{T}|(\mathbf{C} \setminus \{X\}) \cup \mathbf{Z}$ because $X \notin \mathbf{R}(\mathbf{C} \setminus \{X\})$. Then, $Y \perp\!\!\!\perp \mathbf{T}|(\mathbf{C} \setminus \{X\}) \cup \mathbf{Z}$ due to contraction and decomposition. Consequently, $Y \perp\!\!\!\perp \mathbf{T}|(\mathbf{C}\setminus\{X\})\cup\mathbf{Z}$ for all $\mathbf{Z} \subseteq \mathbf{U}\setminus\mathbf{T}\setminus(\mathbf{C}\setminus\{X\})\setminus\{Y\}$ and, thus, $Y \notin \mathbf{R}(\mathbf{C} \setminus \{X\})$. When $\mathbf{U} \setminus \mathbf{I}(\mathbf{C})$ is the optimal domain, $\mathbf{U} \setminus \mathbf{I}(\mathbf{C}) \subseteq \mathbf{U} \setminus \mathbf{I}$ which proves that context nodes can help to reduce $\mathbf{U} \setminus \mathbf{I}$.

Finally, it is worth mentioning that Theorems 1-4, which prove that the corresponding methods to identify $\mathbf{R}$ are correct if the independence tests are correct, also prove that the methods are consistent if the tests are consistent, since the number of tests that the methods perform is finite. Kernel-based independence tests that are consistent for any probability distribution exist (Gretton et al. 2005a, 2005b). For Gaussian distributions, the most commonly used independence test is Fisher's $z$ test which is consistent as well (Kalish and Bühlmann 2005). Specifically, these papers show that the probability of error for these tests decays exponentially to zero when the sample size goes to infinity.

## 4 DISCUSSION ON THE ASSUMPTIONS

We now argue that the assumptions of strict positivity, composition and weak transitivity made in Theorems 2-4 are not too restrictive. The assumption of strict positivity is justified in most real-world applications because they typically involve measurement noise.[2] We note that Gaussian distributions are strictly positive. We recall from section 2 that DAG-faithful and UG-faithful probability distributions satisfy composition and weak transitivity. Gaussian distributions satisfy composition and weak transitivity too (Studený 2005). These are important families of probability distributions. The following theorem, which extends Proposition 1 in Chickering and Meek (2002), shows that composition and weak transitivity are conserved when hidden nodes and selection bias exist. For instance, the probability distribution that results from hiding some nodes and instantiating some others in a DAG-faithful probability distribution may not be DAG-faithful but satisfies composition and weak transitivity. This is an important result for gene expression data analysis (see section 6).

---

[2]Note that the fact that the learning data are sparse does not imply that the assumption of strict positivity does not hold. We cannot conclude from a finite sample that some combinations of states are impossible.

**Theorem 5** *Let $p$ be a probability distribution satisfying composition and weak transitivity. Then, $p(\mathbf{U} \setminus \mathbf{W})$ satisfies composition and weak transitivity. Moreover, if $p(\mathbf{U} \setminus \mathbf{W}|\mathbf{W} = \mathbf{w})$ has the same independencies for all $\mathbf{w}$, then $p(\mathbf{U} \setminus \mathbf{W}|\mathbf{W} = \mathbf{w})$ for any $\mathbf{w}$ satisfies composition and weak transitivity.*

**Proof:** Let $\mathbf{X}$, $\mathbf{Y}$ and $\mathbf{Z}$ denote three mutually disjoint subsets of $\mathbf{U} \setminus \mathbf{W}$. Then, $\mathbf{X} \perp\!\!\!\perp \mathbf{Y}|\mathbf{Z}$ in $p(\mathbf{U} \setminus \mathbf{W})$ iff $\mathbf{X} \perp\!\!\!\perp \mathbf{Y}|\mathbf{Z}$ in $p$ and, thus, $p(\mathbf{U} \setminus \mathbf{W})$ satisfies the composition and weak transitivity properties because $p$ satisfies them. Moreover, if $p(\mathbf{U} \setminus \mathbf{W}|\mathbf{W} = \mathbf{w})$ has the same independencies for all $\mathbf{w}$ then, for any $\mathbf{w}$, $\mathbf{X} \perp\!\!\!\perp \mathbf{Y}|\mathbf{Z}$ in $p(\mathbf{U} \setminus \mathbf{W}|\mathbf{W} = \mathbf{w})$ iff $\mathbf{X} \perp\!\!\!\perp \mathbf{Y}|(\mathbf{Z} \cup \mathbf{W})$ in $p$. Then, $p(\mathbf{U} \setminus \mathbf{W}|\mathbf{W} = \mathbf{w})$ for any $\mathbf{w}$ satisfies the composition and weak transitivity properties because $p$ satisfies them. □

If we are not interested in all the queries about $\mathbf{T}$ but only in a subset of them, then it seems reasonable to remove from $\mathbf{U}$ all the nodes that do not take part in any of the queries of interest before starting the analysis of section 3. Let $\mathbf{W}$ denote the nodes removed. The theorem above guarantees that $p(\mathbf{U} \setminus \mathbf{W})$ satisfies the assumptions for the analysis of section 3 if $p$ satisfies them.

In the theorem above, we assume that $p(\mathbf{U} \setminus \mathbf{W}|\mathbf{W} = \mathbf{w})$ has the same independencies for all $\mathbf{w}$. Although such an assumption is not made in Proposition 1 in Chickering and Meek (2002), the authors agree that it is necessary for the proposition to be correct (personal communication). Without the assumption, there can exist context-specific independencies that violate composition or weak transitivity. An example follows. Let $p(X, Y, Z, W)$ be the probability distribution represented by a BN with four binary nodes, structure $\{Pa(X) = Pa(Y) = Pa(W) = \emptyset, Pa(Z) = \{X, Y, W\}\}$, and parameters $p(X) = p(Y) = p(W) = (0.5, 0.5)$, $p(Z|X, Y, W = 0) = XOR(X, Y)$ and $p(Z|X, Y, W = 1) = OR(X, Y)$. Then, $p(X, Y, Z, W)$ is faithful to the BN structure and, thus, satisfies composition while $p(X, Y, Z|W = 0)$ does not.

We now argue that it is not too restrictive to assume, like in Theorems 4 and 5, that $p(\mathbf{U} \setminus \mathbf{W}|\mathbf{W} = \mathbf{w})$ has the same independencies for all $\mathbf{w}$. Specifically, we show that there are important families of probability distributions where such context-specific independencies do not exist or are very rare. If $p$ is a Gaussian distribution, then $p(\mathbf{U} \setminus \mathbf{W}|\mathbf{W} = \mathbf{w})$ has the same independencies for all $\mathbf{w}$, because the independencies in $p(\mathbf{U} \setminus \mathbf{W}|\mathbf{W} = \mathbf{w})$ only depend on the variance-covariance matrix of $p$ (Anderson 1984). Let us now focus on all the multinomial distributions $p$ for which a DAG $G$ is an independence map and denote them by $M(G)$. The following theorem, which is inspired by Theorem 7 in Meek (1995), shows that the probability of randomly drawing from $M(G)$ a probability distribution with context-specific independencies is zero.

**Theorem 6** *The probability distributions $p$ in $M(G)$ for which there exists some $\mathbf{W} \subseteq \mathbf{U}$ such that $p(\mathbf{U} \setminus \mathbf{W}|\mathbf{W} = \mathbf{w})$ does not have the same independencies for all $\mathbf{w}$ have Lebesgue measure zero wrt $M(G)$.*

**Proof:** The proof basically proceeds in the same way as that of Theorem 7 in Meek (1995), so we refer the reader to that paper for more details. Let $\mathbf{W} \subseteq \mathbf{U}$. Let $\mathbf{X}$, $\mathbf{Y}$ and $\mathbf{Z}$ denote three disjoint subsets of $\mathbf{U} \setminus \mathbf{W}$. For a constraint such as $\mathbf{X} \perp\!\!\!\perp \mathbf{Y}|\mathbf{Z}$ to be true in $p(\mathbf{U} \setminus \mathbf{W}|\mathbf{W} = \mathbf{w})$ but false in $p(\mathbf{U} \setminus \mathbf{W}|\mathbf{W} = \mathbf{w}')$, the following equations must be satisfied: $p(\mathbf{X} = \mathbf{x}, \mathbf{Y} = \mathbf{y}, \mathbf{Z} = \mathbf{z}, \mathbf{W} = \mathbf{w})p(\mathbf{Z} = \mathbf{z}, \mathbf{W} = \mathbf{w}) - p(\mathbf{X} = \mathbf{x}, \mathbf{Z} = \mathbf{z}, \mathbf{W} = \mathbf{w})p(\mathbf{Y} = \mathbf{y}, \mathbf{Z} = \mathbf{z}, \mathbf{W} = \mathbf{w}) = 0$ for all $\mathbf{x}$, $\mathbf{y}$ and $\mathbf{z}$. Each equation is a polynomial in the BN parameters corresponding to $G$, because each term $p(\mathbf{V} = \mathbf{v})$ in the equations is the summation of products of BN parameters (Meek 1995). Furthermore, each polynomial is non-trivial, i.e. not all the values of the BN parameters corresponding to $G$ are solutions to the polynomial. To see it, it suffices to rename $\mathbf{w}$ to $\mathbf{w}'$ and $\mathbf{w}'$ to $\mathbf{w}$ because, originally, $\mathbf{X} \not\perp\!\!\!\perp \mathbf{Y}|\mathbf{Z}$ in $p(\mathbf{U} \setminus \mathbf{W}|\mathbf{W} = \mathbf{w}')$. Let $sol(\mathbf{x}, \mathbf{y}, \mathbf{z}, \mathbf{w})$ denote the set of solutions to the polynomial for $\mathbf{x}$, $\mathbf{y}$ and $\mathbf{z}$. Then, $sol(\mathbf{x}, \mathbf{y}, \mathbf{z}, \mathbf{w})$ has Lebesgue measure zero wrt $\mathbb{R}^n$, where $n$ is the number of linearly independent BN parameters corresponding to $G$, because it consists of the solutions to a non-trivial polynomial (Okamoto 1973). Then, $sol = \bigcup_{\mathbf{X},\mathbf{Y},\mathbf{Z},\mathbf{W}} \bigcup_{\mathbf{w}} \bigcap_{\mathbf{x},\mathbf{y},\mathbf{z}} sol(\mathbf{x}, \mathbf{y}, \mathbf{z}, \mathbf{w})$ has Lebesgue measure zero wrt $\mathbb{R}^n$ because the finite union and intersection of sets of Lebesgue measure zero has Lebesgue measure zero too. Consequently, the probability distributions in $M(G)$ with context-specific independencies have Lebesgue measure zero wrt $\mathbb{R}^n$ because they are contained in $sol$. Finally, since $M(G)$ has positive Lebesgue measure wrt $\mathbb{R}^n$ (Meek 1995), the probability distributions in $M(G)$ with context-specific independencies have Lebesgue measure zero wrt $M(G)$. □

## 5   AN EXAMPLE

In this section, we illustrate the method to identify $\mathbf{R}$ that follows from Theorems 3 and 4 with an example that includes selection bias and context nodes. Let $p(C_1, C_2, I, S, T)$ be any Gaussian distribution that is faithful to the DAG (a) in figure 1. Such probability distributions exist (Meek 1995). Let us consider the selection bias $S = s$. Then, $p(C_1, C_2, I, T|S = s)$ is faithful to the UG (b) in figure 1 and, thus, is not DAG-faithful (Chickering and Meek 2002). Let us now assume that we want to learn a BN from $p(C_1, C_2, I, T|S = s)$ to answer any query about $T$ with context nodes $\{C_1, C_2\}$. Since $p(C_1, C_2, I, T|S = s)$ is a Gaussian distribution (Anderson 1984), it satisfies the assumptions in Theorem 4. Therefore, we can apply the method that follows from this theorem to identify $\mathbf{R}(\{C_1, C_2\})$. The result is $\mathbf{R}(\{C_1, C_2\}) = \emptyset$ and, thus, $\mathbf{I}(\{C_1, C_2\}) = \{I\}$. Moreover, $\{C_1, C_2, T\}$ is the optimal domain to learn the BN because $C_1 \in \mathbf{R}(\{C_2\}) = \{C_1, I\}$ and $C_2 \in \mathbf{R}(\{C_1\}) = \{C_2, I\}$. So, we have solved the problem correctly without learning a BN first.

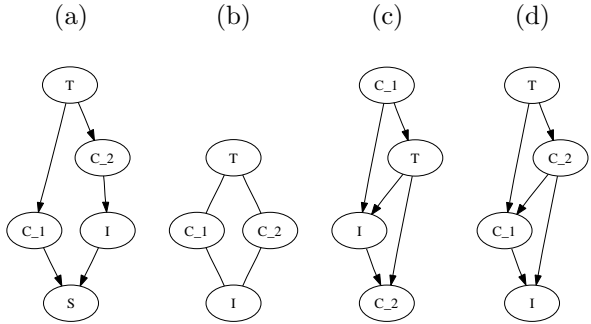

Figure 1: (a) DAG faithful before selection bias, (b) UG faithful after selection bias, (c, d) the only minimal directed independence maps after selection bias.

In the cases where a BN can be learnt first, it is tempting to try to identify $\mathbf{R}$ by just studying the BN structure learnt. This can however lead to erroneous conclusions. For instance, the best BN structure that can be learnt from $p(C_1, C_2, I, T|S = s)$ is a minimal directed independence map of it because, as discussed above, this probability distribution is not DAG-faithful. The DAGs (c) and (d) in figure 1 are the only minimal directed independence maps of $p(C_1, C_2, I, T|S = s)$ (Chickering and Meek 2002). Let us assume that the BN structure learnt is the DAG (c). Then, it seems reasonable to conclude that $\mathbf{R}(\{C_1, C_2\}) = \{I\}$, because $sep(I, T|\{C1, C_2\})$ is false in the BN structure learnt and there exist probability distributions that are faithful to it (Meek 1995). In other words, by just studying the BN structure learnt, it is not possible to detect whether we are dealing with a probability distribution that is faithful to it or not and, thus, it is safer to declare $I$ relevant. We know that this is not the correct solution. The same problem occurs with $C_1$ in the DAG (d) in figure 1 if we want to answer any query about $C_2$ with context nodes $\{I, T\}$. As the authors note, the algorithms in Geiger et al. (1990) and Shachter (1988, 1990, 1998) also suffer from this drawback. On the other hand, our method avoids it by studying the probability distribution instead of a possibly inaccurate BN structure learnt from it.

## 6 CONCLUSIONS

We have reported a method to identify all the nodes that are relevant to compute all the conditional probability distributions for a given set of nodes without having to learn a BN first. We have showed that the method is efficient and consistent under some assumptions. We have argued that the assumptions are not too restrictive. For instance, composition and weak transitivity, which are the two main assumptions, are weaker than faithfulness. We believe that our work can be helpful to those dealing with high-dimensional domains. This paper builds on the fact that the minimal undirected independence map of a strictly positive probability distribution that satisfies weak transitivity can be used to read certain dependencies (Theorem 2). In (Peña et al. 2006b), we introduce a sound and complete graphical criterion for this purpose.

We are currently studying even less restrictive assumptions than those in this paper. The objective is develop a new method whose assumptions are satisfied by such an important family of probability distributions as the family of conditional Gaussian distributions because, in general, this family does not satisfy composition. An example follows. Let $\{X, Y, Z\}$ be a random variable such that $X$ and $Y$ are continuous and $Z$ is binary. Let $p(X, Y|Z = z)$ be a Gaussian distribution for all $z$. Let these two Gaussian distributions have the same mean and diagonal of the variance-covariance matrix but different off-diagonal. Then, $\{X, Y\} \not\perp Z|\emptyset$ but $X \perp Z|\emptyset$ and $Y \perp Z|\emptyset$ (Anderson 1984). We note that there do exist conditional Gaussian distributions that satisfy composition and weak transitivity, e.g. those that are DAG-faithful.

We are also currently applying the method proposed in this paper to our atherosclerosis gene expression database. We believe that it is not unrealistic to assume that the probability distribution underlying our data satisfies strict positivity, composition and weak transitivity. The cell is the functional unit of all the organisms and includes all the information necessary to regulate its function. This information is encoded in the DNA of the cell, which is divided into a set of genes, each coding for one or more proteins. Proteins are required for practically all the functions in the cell. The amount of protein produced depends on the expression level of the coding gene which, in turn, depends on the amount of proteins produced by other genes. Therefore, a dynamic BN is a rather accurate model of the cell (Murphy and Mian 1999): The nodes represent the genes and proteins, and the edges and parameters represent the causal relations between the gene expression levels and the protein amounts. It is important that the BN is dynamic because a gene can regulate some of its regulators and even itself with some time delay. Since the technology for measuring the state of the protein nodes is not widely available yet, the data in most projects on gene expression data analysis are a sample of the probability distribution represented by the dynamic Bayesian network after hiding the protein nodes. The probability distribution with no node hidden is almost surely faithful to the dynamic Bayesian network (Meek 1995) and, thus, it satisfies composition and weak transitivity (see section 2) and, thus, so does the probability distribution after hiding the protein nodes (see Theorem 5). The assumption that the probability distribution sampled is strictly positive is justified because measuring the state of the gene nodes involves a series of complex wet-lab and computer-assisted steps that introduces noise in the measurements (Sebastiani et al. 2003).

**Acknowledgements**

We thank the anonymous referees for their comments and pointers to relevant literature. We thank D. M. Chickering and C. Meek for kindly answering our questions about Proposition 1 in Chickering and Meek (2002). This work is funded by the Swedish Research Council (VR-621-2005-4202), the Swedish Foundation for Strategic Research, and Linköping University.

**References**

Anderson, T. W.: An Introduction to Multivariate Statistical Analysis. John Wiley & Sons (1984).

Chickering D. M., Meek C.: Finding Optimal Bayesian Networks. In Proceedings of the Eighteenth Conference on Uncertainty in Artificial Intelligence (2002) 94-102.

Geiger, D., Verma, T., Pearl, J.: Identifying Independence in Bayesian Networks. Networks 20 (1990) 507-534.

Gretton, A., Herbrich, R., Smola, A., Bousquet, O., Schölkopf, B.: Kernel Methods for Measuring Independence. Journal of Machine Learning Research 6 (2005a) 2075-2129.

Gretton, A., Smola, A., Bousquet, O., Herbrich, R., Belitski, A., Augath, M., Murayama, Y., Pauls, J., Schölkopf, B., Logothetis, N.: Kernel Constrained Covariance for Dependence Measurement. In Proceedings of the Tenth International Workshop on Artificial Intelligence and Statistics (2005b) 1-8.

Kalisch, M., Bühlmann, P.: Estimating High-Dimensional Directed Acyclic Graphs with the PC-Algorithm. Technical report (2005). Available at http://stat.ethz.ch/~buhlmann/bibliog.html.

Lin, Y., Druzdzel, M. J.: Computational Advantages of Relevance Reasoning in Bayesian Belief Networks. In Proceedings of the Thirteenth Conference on Uncertainty in Artificial Intelligence (1997) 342-350.

Madsen, A. L., Jensen, F. V.: Lazy Propagation in Junction Trees. In Proceedings of the Fourteenth Conference on Uncertainty in Artificial Intelligence (1998) 362-369.

Mahoney, S. M., Laskey, K. B.: Constructing Situation Specific Belief Networks. In Proceedings of the Fourteenth Conference on Uncertainty in Artificial Intelligence (1998) 370-378.

Meek, C.: Strong Completeness and Faithfulness in Bayesian Networks. In Proceedings of the Eleventh Conference on Uncertainty in Artificial Intelligence (1995) 411-418.

Murphy, K., Mian, S.: Modelling Gene Expression Data Using Dynamic Bayesian Networks. Technical report (1999). Available at http://www.cs.ubc.ca/~murphyk/papers.html.

Okamoto, M.: Distinctness of the Eigenvalues of a Quadratic Form in a Multivariate Sample. Annals of Statistics 1 (1973) 763-765.

Pearl, J.: Probabilistic Reasoning in Intelligent Systems: Networks of Plausible Inference. Morgan Kaufmann (1988).

Peña, J. M., Nilsson, R., Björkegren, J., Tegnér, J.: Towards Scalable and Data Efficient Learning of Markov Boundaries. International Journal of Approximate Reasoning (2006a) submitted. Available at http://www.ifm.liu.se/~jmp/ijarecsqarujmp.pdf.

Peña, J. M., Nilsson, R., Björkegren, J., Tegnér, J.: Reading Dependencies from the Minimal Undirected Independence Map of a Graphoid that Satisfies Weak Transitivity. The Third European Workshop on Probabilistic Graphical Models (2006b) submitted. Available at http://www.ifm.liu.se/~jmp/jmppgm2006.pdf.

Sebastiani, P., Gussoni, E., Kohane, I. S., Ramoni, M.: Statistical Challenges in Functional Genomics (with Discussion). Statistical Science 18 (2003) 33-60.

Shachter, R. D.: Probabilistic Inference and Influence

Diagrams. Operations Research 36 (1988) 589-604.

Shachter, R. D.: An Ordered Examination of Influence Diagrams. Networks 20 (1990) 535-563.

Shachter, R. D.: Bayes-Ball: The Rational Pastime (for Determining Irrelevance and Requisite Information in Belief Networks and Influence Diagrams). In Proceedings of the Fourteenth Conference on Uncertainty in Artificial Intelligence (1998) 480-487.

Studený, M.: Probabilistic Conditional Independence Structures. Springer (2005).

Tsamardinos, I., Aliferis, C. F., Statnikov, A.: Algorithms for Large Scale Markov Blanket Discovery. In Proceedings of the Sixteenth International Florida Artificial Intelligence Research Society Conference (2003) 376-380.